\documentclass[]{interact}

\usepackage{epstopdf}
\usepackage[caption=false]{subfig}

\usepackage[numbers,sort&compress]{natbib}
\bibpunct[, ]{[}{]}{,}{n}{,}{,}
\renewcommand\bibfont{\fontsize{10}{12}\selectfont}

\theoremstyle{plain}

\theoremstyle{definition}

\theoremstyle{remark}

\usepackage[table]{xcolor}

\begin{document}

\articletype{ORIGINAL ARTICLE}

\title{Computer vision application for improved product traceability in the granite manufacturing industry}

\author{
\name{Xurxo Rigueira \textsuperscript{a}, Javier Martinez \textsuperscript{b}\thanks{CONTACT Javier Martinez. Email: javmartinez@uvigo.es}, Maria Araujo \textsuperscript{a} and Antonio Recaman \textsuperscript{c}}
\affil{\textsuperscript{a}Department of Natural Resources and Environmental Engineering, University of Vigo, Vigo, Spain; \textsuperscript{b}Department of Applied Mathematics I, University of Vigo, Vigo, Spain; \textsuperscript{c}Pavestone S.L., Carretera el Berrueco (km 1), La Cabrera, Madrid, Spain}
}

\maketitle

\begin{abstract}
The traceability of granite blocks consists in identifying each block with a finite number of color bands which represent a numerical code. This code has to be read several times throughout the manufacturing process, but its accuracy is subject to human errors, leading to cause faults in the traceability system. A computer vision system is presented to address this problem through color detection and the decryption of the associated code. The system developed makes use of color space transformations, and several thresholds for the isolation of the colors. Computer vision methods are implemented, along with contour detection procedures for color identification. Lastly, the analysis of geometrical features is used to decrypt the color code captured. The proposed algorithm is trained on a set of 109 pictures taken in different environmental conditions and validated on a set of 21 images. The outcome shows promising results with an accuracy rate of 75.00\% in the validation process. Therefore, the application presented can help employees reduce the number of mistakes on product tracking.
\end{abstract}


\section{Introduction}
\label{intro}

The granite industry holds a key role in the industrial network of the North-Western area of Spain. This region is the largest ornamental granite producer, accounting for 62\% of the total national production \citep{estadisticaminera2019}. Traditionally the mining sector, including the granite industry, has been reluctant to adopt new technologies, but the competitive market leaves few options for the mining industry \citep{Qi2020}.

In this context, an improvement in traceability is required. Traceability, by definition, involves any processes, procedures, or systems that support the generation of verifiable evidence about a product as it moves along its supply chain \citep{AnhVo2018}. Granite blocks complicate traceability tasks due to their broad range of mineralogical characteristics and origins. Nowadays, blocks are initially identified in the quarry with marks indicating type and origin, once they reach the factory for further processing, they are color-coded for identification purposes \citep{Araujo2010}. Moreover, the code is associated through a database with the technical information of each granite block including its variety, quality, supplier, and origin.

Despite this method being the cheapest available for the industry, it entails a substantial number of drawbacks, which may lead to failures. The colors have to be identified in several steps of the production process by an employee. This simple task can be subject to human error due to fatigue and the harsh work environment in the granite industry. On top of that, the granite slabs tend to be stocked on the outside for long periods between 1 and 8 weeks, which gives enough time for the weathering of the colors adding more difficulty for the workers.

A computer vision application can prove helpful to solve this problem and update the traceability system in use. The scope of computer vision is the development of theories and algorithms for automating the process of visual perception. Its applications reach a broad range of fields and the potential benefit of its development can have a major impact in the upcoming years. Computer vision has found its applications in the mining sector, more specifically, in the slate industry for the detection of defects \citep{Iglesias2018}, \citep{Martinez2013} and \citep{Lopez2010Sensors}, in the ceramic industry for the same purpose \citep{Ozkan2016}, \citep{Hanzaei2017}, \citep{Sioma2020}, \citep{Hocenski2016} and \citep{Samarawickrama2017}, in the marble industry for pattern detection and classification \citep{Avci2021} and \citep{Satapathy2018} as well as in the granite industry for the characterization of granite varieties \citep{Araujo2010} and \citep{Lopez2010Mathematics}, but there has not been found any previous research on the applications of computer vision for color detection and analysis in order to improve product traceability in this sector.

The main goal of the program developed is to analyze pictures of granite slabs with a color code drawn on their side and convert the color bands into a numerical code. This task is currently performed manually, therefore, the implementation of a program such as the one presented would dramatically reduce the time needed to transform the color code into its numeric equivalent and reduce human error. These two benefits improve the traceability of the granite slabs throughout the manufacturing process because it makes the identification process more simple and reliable. In addition, it reduces economic losses because those slabs which can not be accurately identified are immediately discarded, which counts up as environmental pollution. The work presented in this research paper is a crucial step in the implementation of this system on cellphone devices for in-factory use with the same purpose and even more features.

Section 2 of this paper introduces the materials needed for this research, including the granite slabs, the images with the color bands, a middle-class computer, and Python 3.9. Section 3 explains the methodology of this research focused on the algorithm developed, while Section 4 presents the results obtained on the detection of color bands on granite and the decryption process. Lastly, Section 5 concludes with the main findings and future applications.

\section{Materials} 
\label{materials}
Granite is by definition a very hard natural igneous rock formation of visibly crystalline texture formed essentially of quartz and orthoclase or microcline and used especially for building and monuments. It typically contains 20-60\% quartz, 10-65\% feldspar, and 5-15\% mica (biotite or muscovite), although a wide range of lithological materials is considered granite from a commercial perspective \citep{Araujo2010}. Granite blocks mined in the quarry are wired cut when processed in the factory to produce the granite slabs. Their dimensions are not consistent and tend to range between 1.70 to 2.00 meters in width and 20 to 30 mm in thickness.

The database used in this research project consists of 130 pictures of granite slabs focused on one of the sides of the slab, where the color code displayed is in the shape of bands manually drawn. The sole purpose of this particular type of code is to keep track of the granite slab from its entry into the factory until its final sale to the customer.

The color code is made up of a variable number of color bands drawn using spray paint on the granite block once it enters the factory. They are placed either close to the higher or lower edges of the block, with a distance of approximately 15 cm in between them. Each band can have any of the 8 different colors shown in Table \ref{tab:Table_1}, and they can be placed in any order. Each color is associated with a number, which makes the block have a specific numerical code depending on the configuration of the color bands it has drawn on its side. This method to numerate, count and keep track of every block has been chosen over simply drawing the corresponding number on the block because, to produce the final version of the product, the block has to be wired cut. This means that the color bands drawn on the surface of the stone would be cut as well, and they would become illegible, losing their purpose. 

\begin{table}
    \centering
    \normalsize
    \tbl{Code key for decoding the encrypted unique number that corresponds to every individual granite block processed in the factory. Each color is associated to an integer between 0 and 7. The number of color bands displayed grows depending on the amount of blocks that the factory has processed. The images in the database display five bands, which corresponds with positive numbers of five digits.}
    {\begin{tabular}{*{10}{c}}
    \hline
    \cellcolor[HTML]{000000} \textcolor{white}{Black} & \cellcolor[HTML]{663300} \textcolor{white}{Brown} & \cellcolor[HTML]{FF0000} Red & \cellcolor[HTML]{FFA500} Orange & \cellcolor[HTML]{FFFF00} Yellow & \cellcolor[HTML]{008000} Green & \cellcolor[HTML]{0000FF} \textcolor{white}{Blue} & \cellcolor[HTML]{800080} \textcolor{white}{Purple} \\
    \hline
    0 & 1 & 2 & 3 & 4 & 5 & 6 & 7 \\
    \hline
    \end{tabular}}
    \label{tab:Table_1}
\end{table} 

The camera used to capture the side of the granite slabs has a resolution of 12MP, a ƒ/1.6 aperture, 26 mm of focal length. Development and testing of the program have been completely performed with a 1.6 GHz Intel i5 10210U processor with 16 GB (2×8 GB) of RAM. The program to identify color codes is written in Python 3.9 using OpenCV, which is an open-source computer vision library written in C++ and C.

\section{Methodology}
\label{methodology}

\subsection{Theory of color detection}
\label{basisofcolordeceection}
A correct choice of color space is considered crucial in computer vision applications. There are several types of colors spaces available, but for the most part, they can be classified into two categories: device-dependent or device-independent. The first group includes those color spaces, which their representation of colors lays further away when compared to how the human nervous system senses colors. According to \citep{Kang2006}, the color spaces in this category, such as RGB and HSV, simply encode device-specific data at the device level. On the other hand, the color spaces in the second group are directly related to the human visual system. They aim to define color coordinates, which can be comprehended by the average observer. The basic color space in this category is CIE XYZ and any other color spaced that can be transformed directly into CIE XYZ is considered device-independent, such as CIE Lab or CIE Luv. Moreover, the concept of uniformity can be applied in this category: a color space is said uniform when the Euclidean distance between colors in that space is proportional to color differences as perceived by humans \citep{Bianconi2013}.

Images are captured in the RGB color space, whereas hue-based color models, such as HSV, are the most commonly implemented in color detection in OpenCV due to their robustness against light changes, while CIE Lab can be more efficient measuring color differences in brightness. According to \citep{Paschos2000}, HSV outperforms RGB because it is approximately uniform and divides the color data into intensity (Value) and a chromatic part (Hue and Saturation).

\subsubsection{RGB color space}
\label{RGB}
The RGB abbreviation stands for the names of its three channels: red, green, and blue, while the rest of the colors can be made up by the combination with different intensities of those three primal colors. Therefore, the RGB color space can be represented by the 3-D Cartesian coordinate system in which the three axes, X, Y, and Z, correspond to the three channels, R, G, and B, respectively; in this representation, every point is defined by the three distinct values of brightness corresponding to red, green, and blue between zero and one. 

\begin{figure}
    \centering
    \includegraphics[scale = 0.2]{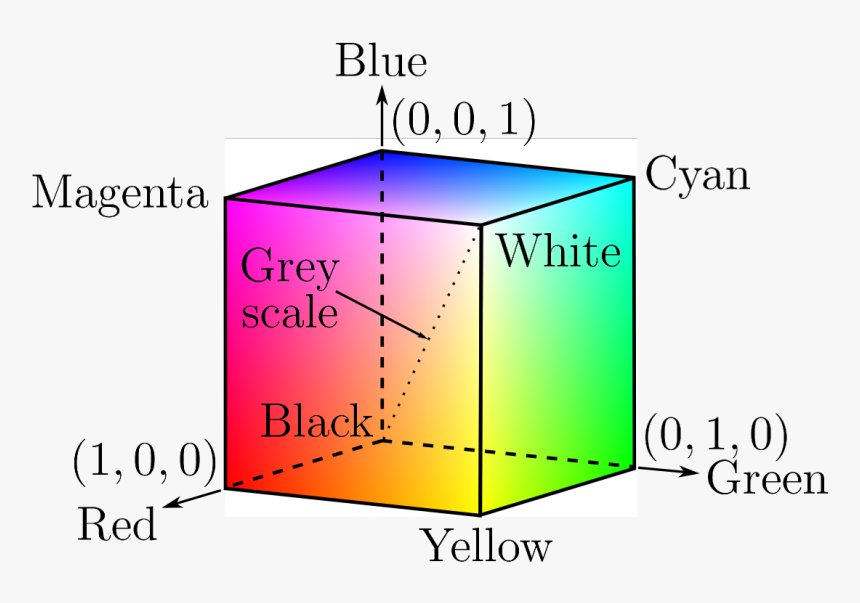}
    \caption{Representation of the RGB color space in Cartesian coordinates by using its three mutually perpendicular axes, where red (R), green (G) and blue (B) are the equivalents of the X, Y, and Z axes.}
    \label{fig:Fig_1}
\end{figure}

This representation can be observed in Figure \ref{fig:Fig_1},  where point (0, 0, 0) is black, (1, 1, 1) is white and the line that joins these two points contains all brightness values for the gray color. The three remaining corners represent the complementary color of the primary colors: yellow, cyan, and magenta \citep{Xiong2018}.

Two of the main advantages of the RGB color space are its simplicity and its additive property, which makes it very easy to store and display images \citep{Ibraheem2012}. Nevertheless, this color model is not the best for color detection, since important color properties such as brightness and purity are embedded within the RGB color channels, which makes it difficult to determine specific colors and their reliable working ranges \citep{Sebastian2010}. Furthermore, the original pictures were processed in the 24-bit RGB format, in which all components have a depth of 8 bits. This makes up a maximum number of colors of $2^8*2^8*2^8=16777216$. The components are internally presented as unsigned integers in range $[0, 255]$, which is the exact value range of a single byte \citep{Chernov2015}.

\subsubsection{HSV color space}
\label{HSV}
In the HSV color space, hue (H) contains the color angle information, saturation (S) represents purity what allows how color is diluted by white. Lastly, value (V) stores the brightness of the color, which measures how far a color is from black. The separation of crucial properties such as brightness and purity is what makes the HSV color space a better fit for color detection purposes.

\begin{figure}
    \centering
    \includegraphics[scale = 0.165]{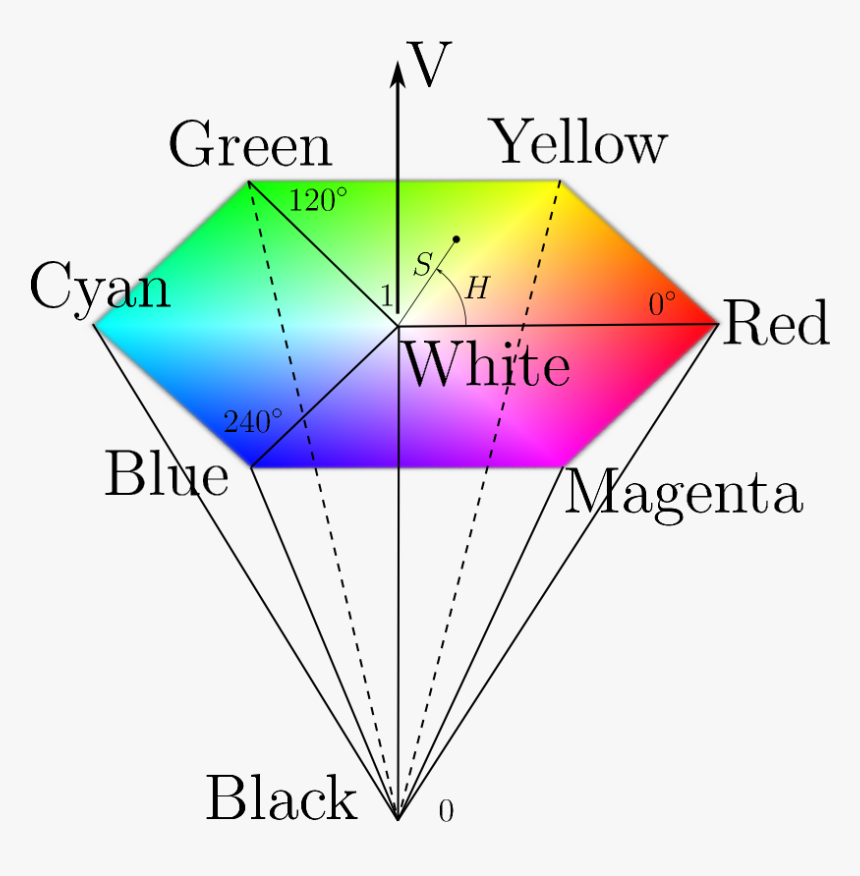}
    \caption{Hexagonal representation of the HSV color space where the central axis contains all the different shades of gray from black to white and all colors can be defined by hue (H), saturation (S) and value (V). Since hue is an angular measure, it becomes highly effective in defining pure colors.}
    \label{fig:Fig_2}
\end{figure}

The HSV color space is a transformation of the RGB color space, and it can be represented in a different coordinate system. The idea of a representation in a hexagonal cone (hexacone) was first proposed by \citep{Smith1978}, and it can be observed in Figure \ref{fig:Fig_2}.

The main diagonal of the cone corresponds to its central axis, which contains all the colors in grayscale from black, at the bottom of the cone, to white, located at the top. The horizontal cross-sections of the hexacone are hexagons of different sizes. As value (V) changes, the sizes of the hexagons are also changed. Each one is larger than the preceding one when moving from black to white. The cross-section that corresponds to black degrades to a single point. 

Points in the hexacone are defined by hue (H), saturation (S), and value (V). Since the hue is an angular value, it can be measured around the main axis of the cone. This helps define the pure colors, which, according to the original layout\citep{Smith1978}, are placed as follows: 0º is red, 60º corresponds to yellow, 120º is green, 180º is cyan, 240º corresponds to blue, and 300º is magenta.

The transformation from RGB to HSV as described in \citep{Rogers1985} is given by \citep{Bhatia2018} in the following equations:

\begin{equation} \label{eq:Eq_1}
\small
V=\max (R, G, B)
\end{equation}
\begin{equation} \label{eq:Eq_2}
\small
S= \begin{cases}\frac{\max (R, G, B)-\min (R, G, B)}{\max (R, G, B)} & \text { if } \max (R, G, B) \neq 0 \\ 0 & \text { otherwise }\end{cases}
\end{equation}
\begin{equation} \label{eq:Eq_3}
\small
H^{\prime}= \begin{cases}\text { undefined } & \text { if } S=0 \\ \frac{G-B}{\max (R, G, B)-\min (R, G, B)} & \text { if } R=\max (R, G, B) \\ 2+\frac{B-R}{\max (R, G, B)-\min (R, G, B)} & \text { if } G=\max (R, G, B) \\ 4+\frac{R-G}{\max (R, G, B)-\min (R, G, B)} & \text { if } B=\max (R, G, B)\end{cases}
\end{equation}
For our purposes, and due to its advantages, the HSV model was chosen to detect the different colors present on the granite slabs.

\subsection{Workflow implemented}
\label{workflowimplemented}

\begin{figure}
    \centering
    \includegraphics[scale=0.095]{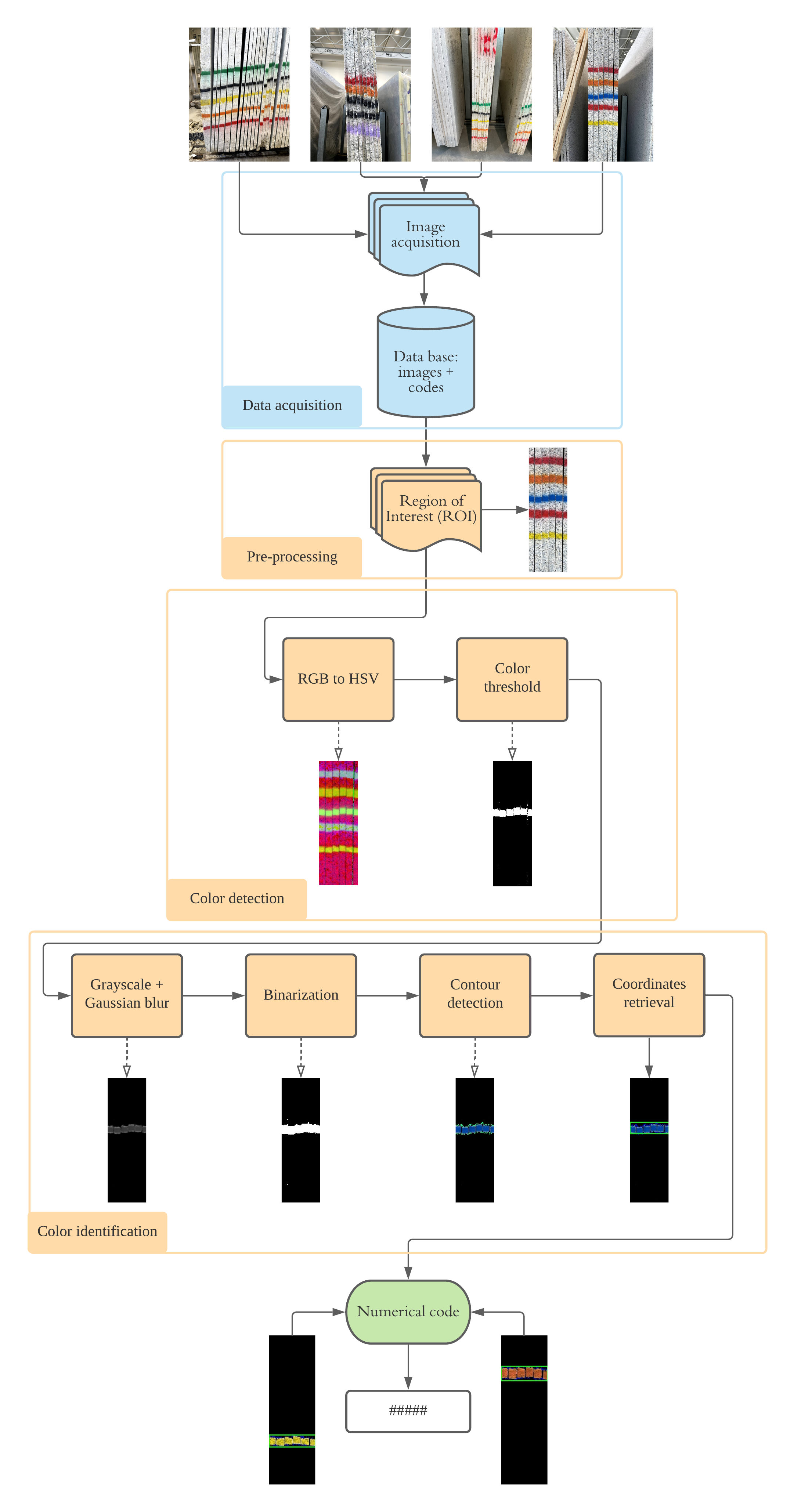}
    \caption{Workflow of the proposed computer vision system with the following steps: (1) data acquisition: the pictures used were taken under different lighting conditions -cloudy and sunny days- as well as varying distances and angles to the side of the granite slab, (2) image pre-processing: images are scaled to a 15\% of their original size and the target area is cropped by the user. (3) Color detection: conversion between the RGB and HSV color spaces and thresholds are implemented to isolate the color bands, (4) color identification: blurring, binarization, and contour detection help retrieve the coordinates of each color band which define the order of the resulting numerical code of the granite slab.}
    \label{fig:Fig_3}
\end{figure}

The presented program uses a computer-vision-based approach to identify different color bands on granite slabs and output a numerical code defined by the color of each band. The workflow diagram of the proposed system is presented in Figure \ref{fig:Fig_3}, where the following stages are introduced:

\begin{enumerate}
    \item Data acquisition.
    \item Image pre-processing.
    \item Color detection.
    \item Color identification.
\end{enumerate}

\subsubsection{Data acquisition}
\label{dataacquisition}
The data were obtained \textit{in situ} under the effect of the changing conditions of the environment and analyzed afterward. As a simple requirement, the color bands must be completely visible. This means that the picture cannot be taken up too close to the point some colors are left out of the image or from a very distant location where the color bands are unidentifiable.

\subsubsection{Image pre-processing}
\label{preprocessing}
The pre-processing of the images includes two main operations. In the first place, pictures are scaled down to 15\% of their initial dimension, decreasing the computational load without compromising the final outcome. Furthermore, the program allows the user to crop out the area where the color bands are captured on the scaled picture, which entails an increase in the overall accuracy of the results obtained. 

\subsubsection{Color detection}
\label{colordetection}
Color detection is performed by a total number of eight functions, one for every color. They share a common algorithm, but the input parameters change depending on the target color. These inputs are not arbitrary, initially, they were set manually, but in order to increase the precision of the model, the parameters have to be optimized for every case. This is done in the training process, which aims to minimize the error on color detection and identification. In order to do this, the algorithm iterates on the available images for every combination possible within the ranges set for each parameter. A success condition is defined, which helps identify those combinations which succeed in their task, therefore, those parameters with a higher number of true positives become the ideal values in each function. 

The input parameters for all functions are (1) color name, (2) associated code number, (3) colored area (CA), (4) colored ratio (CR), (5) width/height ratio (WHR), (6) maximum vertical distance (MVD), (7) minimum Hue (Hmin), (8) minimum Saturation (Smin), (9) minimum Value (Vmin), (10) maximum Hue (Hmax), (11) maximum Saturation (Smax), and (12) maximum Value (Vmax). Parameters (1) and (2) are the decoded and encrypted names of each color respectively, the colored area (3) defines the minimum number of pixels that have to be turned on to be considered a true positive case of color detection, while the color ratio (4) is defined by the relation between the number of pixels colored with the target color and the total amount of pixels; it has the same goal as the colored area but implements a second layer of security to avoid false positives. The width/height ratio (5) analyzes the relation between the width and the height of the color areas detected. Given the fact that the color bands have rectangular shapes, this parameter aims to get rid of all detected areas which do not have the minimum width to height ratios, such as shades or color spots that do not belong to the color code system. The minimum vertical distance (6) quantifies the required separation between different color bands to be considered independent. These bands are drawn directly on the granite block, which is cut afterward to generate the granite slabs, in this process the color bands are also divided into several sections. As a consequence of this division, gaps are generated throughout every color band, with a direct effect on the color detection process. When the picture is taken, these gaps are usually shown as dark spaces that break up the color band, as a result, several colored areas will be identified, but thanks to the implementation of this parameter, as long as the detected areas are within the height range set by the MVD, they will be considered as a single band. Lastly, parameters from 7 to 12 belong to the HSV color space and define the threshold for each color.

\subsubsection{Thresholds}
\label{thresholds}
Each color function takes the RGB image, proceeds with the HSV transformation, and applies the color thresholds on the resulting image. In this case, an adaptive multichannel threshold was implemented to identify color in a more reliable way under the wide variety of lighting conditions presented in this problem. The first of the color masks take the lowest hue, saturation, and value possible for the target color, delimiting the range for these three channels. Therefore, once the mask is applied to the HSV picture, the resulting image is binarized as the pixels that are defined within this range are turned into white (1) and the ones which fall out of this condition are turned off and display black (0). The mathematical principle of this method is introduced in Equation \ref{eq:Eq_4}:

\begin{equation} \label{eq:Eq_4}
\scriptsize
    I = \begin{cases} 1 \text{ if } H \in  [H_{min}, H_{max}] \land S \in [S_{min}, S_{max}] \land  V \in [V_{min}, V_{max}] \\ 0 \text{ otherwise }\end{cases}
\end{equation}

Where $H$ is hue, $S$ saturation, and $V$ value. Similarly, the second mask applies the same principle and methodology, but in this case, it takes the highest values of hue, saturation, and value for the target color. The sum of both masks is applied to the original image, concluding the detection of the target color in every case.

\subsection{Color identification}
\label{coloridentification}
In the case that the resulting image delivered by the color detection algorithm fulfills the requirements established by the CA and CR parameters, the image is proposed for further processing where the colors detected are associated with their respective numbers, and the final code for the granite slab is delivered. This process begins with the detection of the contours of all the isolated colored areas. To achieve this, the picture is converted to grayscale and blurred out with the implementation of Gaussian blur to reduce the sharpness of the structures contained in the picture and increase the efficiency of the contour detection method.

Gaussian blur is classified as a low pass filter because it reduces the high-frequency components of the image. It uses Equation \ref{eq:Eq_5} explained in \citep{Shapiro2001} and \citep{Nixon2019}:

\begin{equation} \label{eq:Eq_5}
\small
G(x, y)=\frac{1}{2 \pi \sigma^{2}} e^{-\frac{x^{2}+y^{2}}{2 \sigma^{2}}}
\end{equation}

Where x is the distance from the origin in the horizontal axis, y is the distance from the origin in the vertical axis, and $\sigma$ is the standard deviation of the Gaussian distribution. Applying Equation \ref{eq:Eq_5} produces a surface in which contours are concentric circles with a Gaussian distribution from the center point. The convolution matrix, built from the values of this distribution, is applied to the original image, where the new value of each pixel is set to a weighted average of the pixels in its surrounding area. The pixels further away from the center (original pixel) receive smaller weights as the distance increases, while the original holds the highest value. This helps preserve boundaries and edges in a better way compared to simpler filters.
Lastly, the blurred image is binarized before applying contour detection, in this step all pixels that are black remain in this state, while the rest are converted into white according to Equation \ref{eq:Eq_6}.

\begin{equation}\label{eq:Eq_6}
\small
dst(x,y) = \begin{cases} 1 \text{ if } src(x,y) \in (0, 1]  \\ 0 \text { otherwise } \end{cases}
\end{equation}

Where $src$ is the source image, and $dst$ is the destination image of the same size and type as the source. 

\begin{figure}
    \centering
    \includegraphics[scale = 0.4]{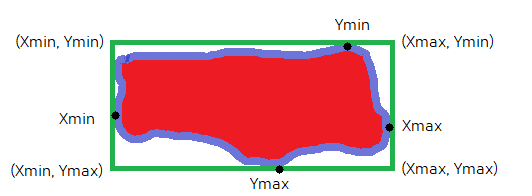}
    \caption{Illustrative representation of how the coordinates of the wrapping rectangle are selected. These are defined by the maximum and minimum values of x and y contained within the perimeter of the colored area.}
    \label{fig:Fig_4}
\end{figure}

The contour detection method \citep{2014opencv} \citep{Suziki1983} is performed on the binarized image. In this process, those areas of the picture with an intensity gradient strong enough to be noticed by the algorithm are detected and marked up with individual points. This produces a point cloud surrounding the different structures featured in the image. A line connects all the significant points around the detected area, and based on the bending of the curve, the algorithm defines where the corners are placed. This allows for the representation of a regular geometric figure, which wraps all around every area detected in this step. Given that the color bands tend to have fairly rectangular shapes, a rectangle was selected for this purpose. Figure \ref{fig:Fig_4} shows a simplified configuration of these rectangles.

\begin{figure}
    \centering
    \subfloat[Cropped.]{%
    \resizebox*{3cm}{!}{\includegraphics{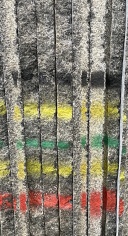}}}\hspace{5pt}
    \subfloat[Green.]{%
    \resizebox*{3cm}{!}{\includegraphics{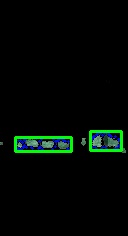}}}
    \caption{Example of a faded green color band in which the algorithm detects two different sections. This scenario could account for a wrong number of bands, but the implementation of the maximum vertical distance (MVD) parameter limits the amount of space between bands to be considered as so and joins under a single band those which have similar \textit{y} coordinates.} 
    \label{fig:Fig_5}
\end{figure}

Up until this last step, there can be colored areas of the picture that fulfill all the requirements to be considered as color bands, but in reality, they are not. Therefore, the parameter WHR is introduced, which sets a limit to the numeric relation between width and height for every rectangle.  It allows filtering out the vast majority of areas with rectangular shapes, eliminating those in which the vertical sides are the longest. Following up comes the implementation of the last parameter (MVD), as it was explained in \ref{colordetection} \textit{Color detection}, its goal is to group under the same band those areas with equal color, but which are broken down in smaller rectangles as seen in Figure \ref{fig:Fig_5}. This step avoids the anomalous detection of a not realistic number of bands. 

\begin{figure}
    \centering
    \includegraphics[scale = 0.18]{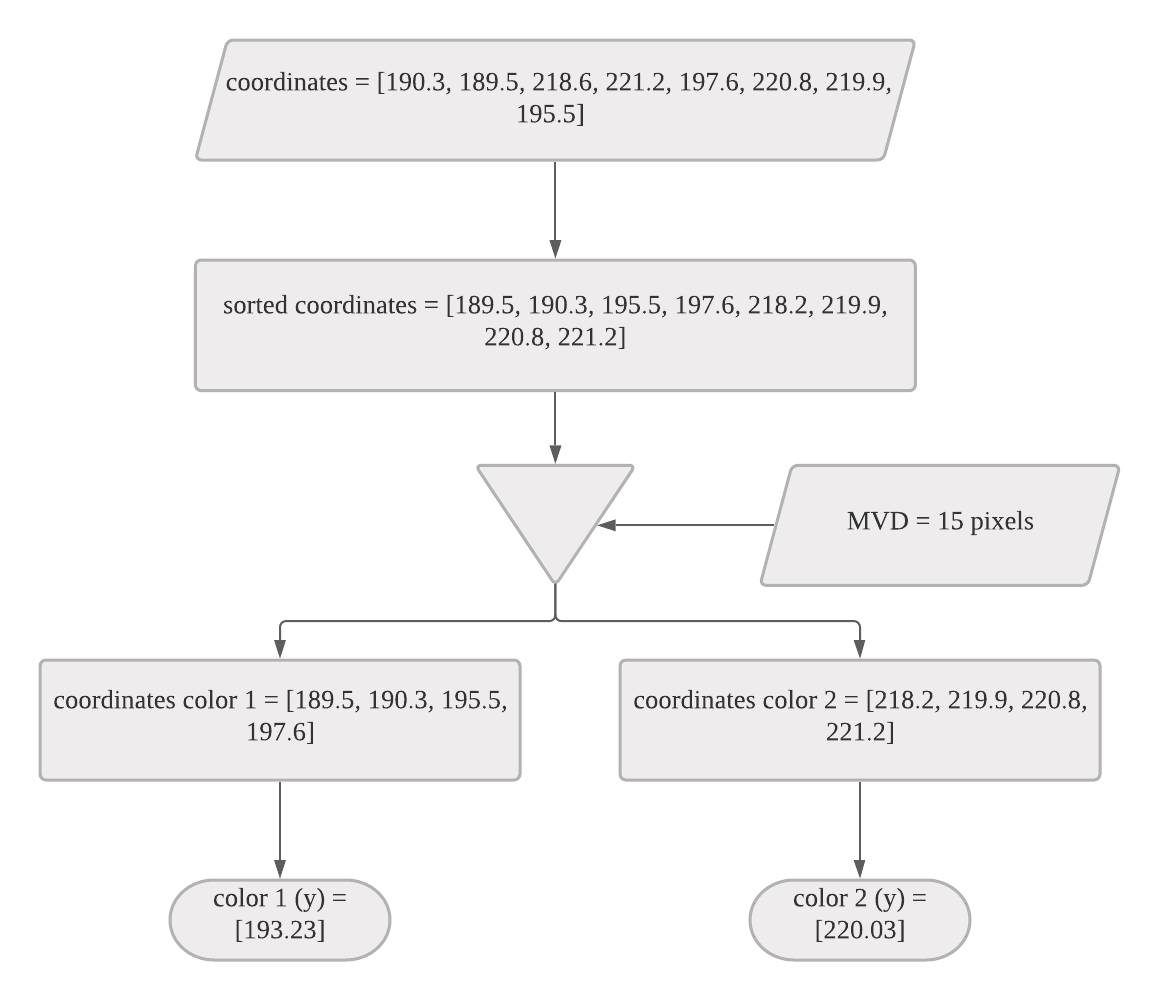}
    \caption{Simplified representation of the process followed by the algorithm to group together detected areas or assign them to separate color bands. Firstly, the coordinates are sorted from least to greatest and put together if the result of their subtraction is smaller than the number defined by the maximum vertical distance parameter, otherwise, they are divided into different color bands.}
    \label{fig:Fig_6}
\end{figure}

The relative coordinates of each band are retrieved from the information contained in the rectangles. For this particular application, the \textit{y} coordinates are sorted from least to greatest and analyzed in this order. In the case the result of the subtraction between $i+1$ and $i$ is smaller than the MVD, those two rectangles would be considered as the same color band, otherwise, they would be set to belong to separate color bands. Lastly, the average of all the coordinates belonging to each color band is calculated to get a single value per band, which allows the algorithm to read the colors and write the code in the correct order. This process is represented in Figure \ref{fig:Fig_6}.

The direction in which the code has to be read is defined by the position of the color bands, meaning that if the bands are placed closer to the top of the slab, the reading direction is downwards, while the reading direction changes to upwards if the bands are placed closer to the bottom of the slab. Essentially, the color code has to be decrypted, starting with the bands which are close to one of the horizontal edges. This distance ratio has to be kept approximately equal when the image is cropped. Therefore, the algorithm can compare which band is closer to its respective reference, top or bottom of the slab, and define the correct reading direction of the code. By knowing which color corresponds to which number, retrieving the coordinates of each color band, and defining the appropriate reading direction, the algorithm can output the encrypted code correctly.

\section{Case study}
\label{casestudy}

The RGB pictures that make up the database of the case studied were taken in three separate batches under non-identical conditions. Their features are shown in Table \ref{tab:Table_2}.

\begin{table}
    \centering
    \tiny
    \tbl{Description of the different batches of pictures that make up the database. The first two columns show the name of the batch and the number of pictures. The next three columns include the distance between the camera and the granite slab, the angle of the camera and the slab, and the area where the picture is focused. The last three columns include the proportion of pictures taken in inside lighting conditions, outside and the resolution of the pictures, respectively.}
    {\begin{tabular}{*{11}{c}}
        \hline
        Name & \# Pictures & Distance(m) & Angle (º) & Focus & \multicolumn{2}{c}{Lighting} & Resolution(pixels) \\
        & & & & & Inside & Outside \\
        \hline
        batch1 & 14 & 1 & Not consistent & Color bands & 100\% & 0.00\% & 3472×4640 \\
        \hline
        batch2 & 40 & 1.75 & Parallel & Color bands & 27.50\% & 72.50\% & 3024×4032 \\
        \hline
        batch3 & 84 & 2 & Parallel & Granite slabs & 25.90\% & 74.10\% & 3024×4032 \\
        \hline
    \end{tabular}}
    \label{tab:Table_2}
\end{table}

The different input parameters mentioned in Section \ref{colordetection} \textit{Color detection} are calculated through the training and validation process of the algorithm for each color function. For this task, the database is divided into two different sets: the training set which has 109 images, and the validation set with 21 images. All functions are trained and validated respectively on those images that have the corresponding target color. The same database structure was used to implement the final algorithm, which is capable of detecting all colors and decrypting the code associated.

Since the color functions need initial parameters to start off the training process, they are given approximate values calculated manually based on the HSV definition of each color for the hue, value, and saturation parameters, while the CA, CR, WHR, and the MVD are iterated from zero to their highest value, being the last one determined from direct analysis of the image features.

\section{Results and discussion}
\label{results}

\subsection{Results by color}
\label{resultsbycolor}

The results obtained for all parameters after the training process are shown in Table \ref{tab:Table_3}.

\begin{table*}
    \centering
    \scriptsize
    \tbl{Final input parameters for each color function obtained by the algorithm after the training process on the 109 images available in the database. The first four parameters are directly related to the geometrical features of the color bands, while the six last columns present the different ranges in the hue, saturation, and value for the detection of each color.}
    {\begin{tabular}{*{11}{c}}
        \hline
        Color & CA & CR & W/HR & MVD & Hmin & Smin & Vmin & Hmax & Smax & Vmax \\
        \hline
        \cellcolor[HTML]{000000} \textcolor{white}{Black} & 50 & 0 & 0.8 & 16 & 0 & 0 & 3 & 179 & 255 & 51 \\
        \hline
        \cellcolor[HTML]{663300} \textcolor{white}{Dark Brown} & 300 & 0 & 1.2 & 72 & 5 & 49 & 64 & 15 & 255 & 128 \\
        \hline
        \cellcolor[HTML]{b35900} Light Brown & 300 & 0 & 1.2 & 72 & 14 & 74 & 132 & 18 & 255 & 165 \\
        \hline
        \cellcolor[HTML]{FF0000} High-Hue Red & 50 & 0 & 0.4 & 22 & 171 & 100 & 103 &179 & 255 & 255 \\
        \hline
        \cellcolor[HTML]{FF0000} Low-Hue Red & 50 & 0 & 0.4 & 30 & 1 & 93 & 97 & 5 & 255 & 255 \\
        \hline
        \cellcolor[HTML]{FFA500} Orange & 250 & 0 & 1.1 & 29 & 6 & 112 & 116 & 15 & 255 & 255 \\
        \hline
        \cellcolor[HTML]{FFFF00} Yellow & 150 & 0 & 0.2 & 16 & 23 & 69 & 42 & 43 & 255 & 255 \\
        \hline
        \cellcolor[HTML]{008000} Green & 150 & 0 & 0.8 & 18 & 40 & 60 & 0 & 80 & 255 & 255 \\
        \hline
        \cellcolor[HTML]{0000ff} \textcolor{white}{Blue} & 100 & 0 & 0.6 & 32 & 81 & 113 & 0 & 109 & 255 & 255 \\
        \hline
        \cellcolor[HTML]{800080} \textcolor{white}{Purple} & 200 & 0 & 0.7 & 30 & 120 & 21 & 81 & 155 & 255 & 255 \\
        \hline
    \end{tabular}}
    \label{tab:Table_3}
\end{table*}

Table \ref{tab:Table_4} includes the conditions and the success rate of every color in the training phase. During the development of the program, it was noticed that the colors red and brown display two different varieties. In order to improve the accuracy, a double mask method was implemented in the algorithm, which works by defining two distinct thresholds for each color variety but still identifying them as the same color. The computational time required for each process is also included.

The changes in the shade of brown have their origin in the weathering of the color. In the case of the red color, we hypothesize that its variance is heavily dependent on the brightness conditions under which the picture was taken. The presence of clouds in the sky or shadows projected on the granite slabs seems to be a likely cause for a high hue value, while if the light rays hit the granite slab and the shadow is projected backward the value of hue for the red color tends to be much lower as it is shown in Table \ref{tab:Table_3}.

\begin{table}
    \centering
    \scriptsize
    \tbl{Results of the training process for each color detection function. The average success rate stands at 84.28\%. The color light brown shows the lowest accuracy due to its similarity with orange.}
    {\begin{tabular}{*{4}{c}}
        \hline
        Color & Images & Success rate & Total time (sec.) \\
        \hline
        \cellcolor[HTML]{000000} \textcolor{white}{Black} & 32 & 90.32\% & 0.37 \\
        \hline
        \cellcolor[HTML]{835C3B} \textcolor{white}{Dark Brown} & 9 & 100.00\% & 0.13 \\
        \hline
        \cellcolor[HTML]{b35900} Light Brown & 32 & 57.69\% & 0.32 \\
        \hline
        \cellcolor[HTML]{FF0000} High-Hue Red & 36 & 83.33\% & 0.29 \\
        \hline
        \cellcolor[HTML]{FF0000} Low-Hue Red & 65 & 79.69\% & 0.54 \\
        \hline
        \cellcolor[HTML]{FFA500} Orange & 57 & 78.95\% & 0.47  \\
        \hline
        \cellcolor[HTML]{FFFF00} Yellow & 60 & 88.33\% & 0.50 \\
        \hline
        \cellcolor[HTML]{008000} Green & 40 & 95.00\% & 0.32 \\
        \hline
        \cellcolor[HTML]{0000ff} \textcolor{white}{Blue} & 32 & 83.97\% & 0.28 \\
        \hline
        \cellcolor[HTML]{800080} \textcolor{white}{Purple} & 38 & 85.53\% & 0.44 \\
        \hline
    \end{tabular}}
    \label{tab:Table_4}
\end{table}

Table \ref{tab:Table_5} presents the results of the validation process by color. It can be seen in the fourth column that the times required for each calculation decrease in direct proportion to the number of images analyzed. The images on this set are completely new for the algorithm, meaning that it has not been trained on them. Therefore, the high success rates in this process confirm that the parameters defined in the training stage are accurate, and the model can read correct codes of new images what makes it valid.

\begin{table}
    \centering
    \scriptsize
    \tbl{Results for the validation process for each color detection function. All colors average a success rate of 86.36\%.}
    {\begin{tabular}{*{4}{c}}
        \hline
        Color & Images & Success rate & Total time (sec.) \\
        \hline
        \cellcolor[HTML]{000000} \textcolor{white}{Black} & 8 & 88.89\% & 0.14 \\
        \hline
        \cellcolor[HTML]{835C3B} \textcolor{white}{Dark Brown} & 4 & 75.00\% & 0.04 \\
        \hline
        \cellcolor[HTML]{b35900} Light Brown & 4 & 75.00\% & 0.04 \\
        \hline
        \cellcolor[HTML]{FF0000} High-Hue Red & 6 & 100.00\% & 0.09 \\
        \hline
        \cellcolor[HTML]{FF0000} Low-Hue Red & 15 & 81.25\% & 0.13 \\
        \hline
        \cellcolor[HTML]{FFA500} Orange & 10 & 80.00\% & 0.09 \\
        \hline
        \cellcolor[HTML]{FFFF00} Yellow & 13 & 85.71\% & 0.16 \\
        \hline
        \cellcolor[HTML]{008000} Green & 9 & 100.00\% & 0.07 \\
        \hline
        \cellcolor[HTML]{0000ff} \textcolor{white}{Blue} & 8 & 88.89\% & 0.09 \\
        \hline
        \cellcolor[HTML]{800080} \textcolor{white}{Purple} & 9 & 88.89\% & 0.09 \\
        \hline
    \end{tabular}}
    \label{tab:Table_5}
\end{table}

\subsection{Combined results}

All parameters for each color have been tuned, and the algorithm is tested on all the images in the training set. The success condition previously defined is changed. In order for a result to be a true hit, the output of the algorithm must match the exact numerical code an employee would read off the granite slab. Those codes obtained from the algorithm whose numbers are not the same and/or are not placed in the right orders would be considered faults.

\begin{figure}
    \centering
    \subfloat[Code 23145.]{%
    \resizebox*{3cm}{!}{\includegraphics{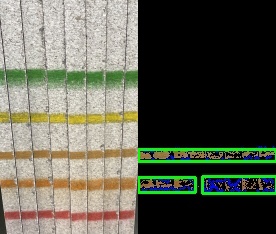}}}\hspace{5pt}
    \subfloat[Code 23823.]{%
    \resizebox*{3cm}{!}{\includegraphics{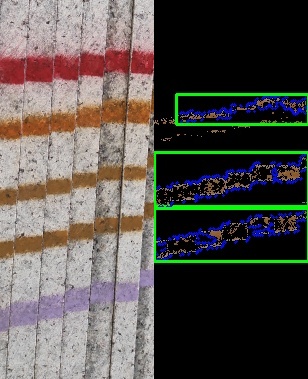}}}
    \caption{Faults in the detection of the color brown in the training set: (a) and (b) failure in the identification of the color brown due to its proximity to orange in the HSV color space.}
    \label{fig:Fig_7}
\end{figure}

Under this condition, the colors which present a lower success rate tend to have an important negative effect on the final result. Particularly, the light brown can be mistaken for orange, and the dark brown can be confused with the low-hue red due to their proximity on the HSV color space. These conditions make it difficult for the algorithm to achieve a very precise detection of all colors, which hijacks the results leading to a faulty detection, as can be seen in Figure \ref{fig:Fig_7}.

\begin{figure}
    \centering
    \subfloat[]{%
    \resizebox*{1cm}{!}{\includegraphics{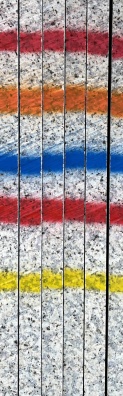}}}\hspace{2pt}
    \subfloat[]{%
    \resizebox*{1cm}{!}{\includegraphics{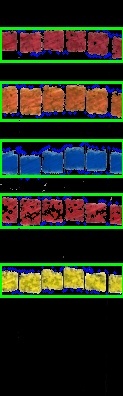}}}\hspace{2pt}
    \subfloat[]{%
    \resizebox*{1cm}{!}{\includegraphics{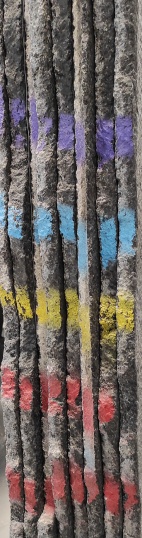}}}\hspace{2pt}
    \subfloat[]{%
    \resizebox*{1cm}{!}{\includegraphics{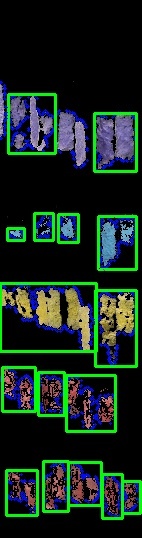}}}\hspace{2pt}
    \subfloat[]{%
    \resizebox*{1cm}{!}{\includegraphics{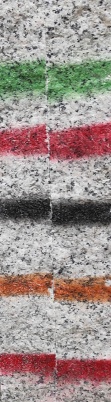}}}\hspace{2pt}
    \subfloat[]{%
    \resizebox*{1cm}{!}{\includegraphics{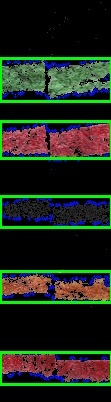}}}\hspace{2pt}
    \subfloat[]{%
    \resizebox*{0.5cm}{!}{\includegraphics{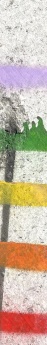}}}\hspace{2pt}
    \subfloat[]{%
    \resizebox*{0.5cm}{!}{\includegraphics{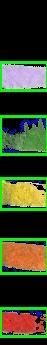}}}
    \caption{Several examples of correct detection of the different color bands and an accurate output of the code represented by those. Images (b) and (e) clearly display the results yielded by the algorithm in the training set, while image (h) belongs to the validation set. Images (c) and (d) show how the algorithm is able to detect color bands that are disrupted in the horizontal axis.}
    \label{fig:Figure_8}
\end{figure}

In order to tackle this issue, the replacement of brown. The spray paint colors cyan and light green are proposed due to their strong contrast with the current colors in use, as shown in Figure \ref{fig:Figure_8}. Moreover, the election of the color pink was motivated by its great endurance to weathering, as it is demonstrated in \citep{Alonso2021}. According to \citep{Alonso2021}, the use of silicate-based paint can be considered as a future line of research to solve the detection problem associated with the erosion of the spray paints currently employed.

Most color detection systems have to operate within steady light conditions \citep{AnhVo2018}, \citep{Iglesias2018}, \citep{Kondo2009}, \citep{Burgos-Artizzu2011} and require stable camera positioning \citep{Carew2003}, \citep{Andrew2016}, \citep{Ghita2005} and \citep{Ghita2006}, while the system proposed, although not perfect, achieves results in very different conditions. In this case, the algorithm shows a success rate of 74.42\% on the training set, while the validation shows efficiency of 75.00\%. Several examples of correct detection are displayed in Figure \ref{fig:Figure_8}.

Table \ref{tab:Table_6} presents the detailed results of the validation process. Here it can be seen how the success condition affects the result, since the numbers have to match the correct code and be placed in the same order. 

\begin{table}
    \centering
    \tiny
    \tbl{Detailed results obtained by the algorithm on the validation set. The success condition turns all incorrect and/or out-of-place digits into a wrong detection.}
    {\begin{tabular}{*{7}{c}}
        \hline
        Code & Number 1 & Number 2 & Number 3 & Number 4 & Number 5 & Result \\
        \hline
        22475 & \cellcolor[HTML]{ffffff} 2 & \cellcolor[HTML]{ffffff} 2 & \cellcolor[HTML]{ffffff} 4 & \cellcolor[HTML]{ffffff} 7 & \cellcolor[HTML]{ffffff} 5 & \cellcolor[HTML]{ffffff} 22475 \\
        \hline
        23405 & \cellcolor[HTML]{ffffff} 2 & \cellcolor[HTML]{ffffff} 3 & \cellcolor[HTML]{ffffff} 4 & \cellcolor[HTML]{ffffff} 0 & \cellcolor[HTML]{ffffff} 5 & \cellcolor[HTML]{ffffff} 23405 \\
        \hline
        23457 & \cellcolor[HTML]{ffffff} 2 & \cellcolor[HTML]{ffffff} 3 & \cellcolor[HTML]{ffffff} 4 & \cellcolor[HTML]{ffffff} 5 & \cellcolor[HTML]{ffffff} 7 & \cellcolor[HTML]{ffffff} 23457 \\
        \hline
        24036 & \cellcolor[HTML]{ffcccc} 4 & \cellcolor[HTML]{ffcccc} 0 & \cellcolor[HTML]{ffffff} 0 & \cellcolor[HTML]{ffffff} 3 & \cellcolor[HTML]{ffffff} 6 & \cellcolor[HTML]{ffcccc} 40036 \\
        \hline
        24066 & \cellcolor[HTML]{ffffff} 2 & \cellcolor[HTML]{ffffff} 4 & \cellcolor[HTML]{ffffff} 0 & \cellcolor[HTML]{ffffff} 6 & \cellcolor[HTML]{ffcccc} 0 & \cellcolor[HTML]{ffcccc} 24060 \\
        \hline
        24227 & \cellcolor[HTML]{ffffff} 2 & \cellcolor[HTML]{ffffff} 4 & \cellcolor[HTML]{ffffff} 2 & \cellcolor[HTML]{ffffff} 2 & \cellcolor[HTML]{ffffff} 7 & \cellcolor[HTML]{ffffff} 24227 \\
        \hline
        24465 & \cellcolor[HTML]{ffffff} 2 & \cellcolor[HTML]{ffffff} 4 & \cellcolor[HTML]{ffffff} 4 & \cellcolor[HTML]{ffffff} 6 & \cellcolor[HTML]{ffffff} 5 & \cellcolor[HTML]{ffffff} 24465 \\
        \hline
        24526 & \cellcolor[HTML]{ffffff} 2 & \cellcolor[HTML]{ffffff} 4 & \cellcolor[HTML]{ffffff} 5 & \cellcolor[HTML]{ffffff} 2 & \cellcolor[HTML]{ffffff} 6 & \cellcolor[HTML]{ffffff} 24526 \\
        \hline
    \end{tabular}}
    \label{tab:Table_6}
\end{table}

\section{Conclusion}
\label{conclusion}

The traceability problem in the granite industry has been studied in the present research paper. The current method to keep track of the granite blocks until they reach the final product stage is fairly simple and cheap, consisting in the usage of colored graffiti spray to assign a number to each granite block. Weathering of the paint and human error due to different causes, such as fatigue, lead to a faulty reading of the numerical code represented by color bands and ultimately to economic losses for the industry as well as pollution originated by the granite that has not been traced. As evidenced by the experimental results shown in this research paper, a color detection algorithm can be developed for future implementation on cellphone devices. The proposed system performs an accurate detection of a combination of seven out of the eight colors used in the granite industry. It shows success rates on the reading of color codes and the output of their corresponding numerical codes of 74.42\% in the training set and 75.00\% in the validation set. Lastly, these results are achieved on images that display different lighting conditions and were not taken with a fixed target distance, what increases the difficulty in the detection of different colors and adds up to the value of the results presented.

\section*{Disclosure statement}

The authors declare that they have no known competing interest financial interest or personal relationships that could have appeared to influence the work reported in this paper

\section*{Funding}

This work was supported by Project PID2020-116013RB-I00 financed by MCIN/AEI/10.13039/501100011033.

\section*{Notes on contributor(s)}

Xurxo Rigueira: Methodology, Software, Investigation, Data Curation, Writing - Original Draft, Visualization; María Araujo: Conceptualization, Validation, Data Curation, Writing - Review \& Editing, Visualization; Javier Martinez: Conceptualization, Methodology, Formal Analysis, Writing - Review \& Editing, Supervision; Eduardo Giraldez: Project Administration, Funding Acquisition; Antonio Recaman: Resources.


\bibliographystyle{tfs.bst}
\bibliography{bibliography.bib}

\end{document}